\documentclass[conference]{IEEEtran}
\usepackage{graphicx,color,multicol,url}

\begin{document}
%
\title{MSCM-LiFe: Multi-scale cross modal linear feature for horizon detection in maritime images}

\author{\IEEEauthorblockN{Dilip~K.~Prasad\IEEEauthorrefmark{1}, Deepu Rajan \IEEEauthorrefmark{2}, C. Krishna Prasath\IEEEauthorrefmark{1},Lily Rachmawati \IEEEauthorrefmark{3}, Eshan Rajabally \IEEEauthorrefmark{4} and Chai Quek \IEEEauthorrefmark{2}}
\IEEEauthorblockA{\IEEEauthorrefmark{1}Rolls-Royce@NTU Corporate Lab, Singapore\\ \IEEEauthorrefmark{2}School of Computer Science and Engineering, Nanyang Technological University, Singapore\\ \IEEEauthorrefmark{3}Rolls-Royce Plc, Singapore\\ \IEEEauthorrefmark{4}Rolls-Royce Derby, United Kingdom}
}



\maketitle
\begin{abstract}
This paper proposes a new method for horizon detection called the multi-scale cross modal linear feature. This method integrates three different concepts related to the presence of horizon in maritime images to increase the accuracy of horizon detection. Specifically it uses the persistence of horizon in multi-scale median filtering, and its detection as a linear feature commonly detected by two different methods, namely the Hough transform of edgemap and the intensity gradient. We demonstrate the performance of the method over 13 videos comprising of more than 3000 frames and show that the proposed method detects horizon with small error in most of the cases, outperforming three state-of-the-art methods.
\end{abstract}

\IEEEpeerreviewmaketitle

\section{Introduction}\label{sec:introduction}
Horizon detection is useful in maritime electro optical data processing for various purposes \cite{Prasad2017Survey}, including registration for mobile sensor platforms such as buoys, maritime vessels, unmanned aerial vehicles \cite{fefilatyev2010tracking,cao2008automatic} and restricting the object search region \cite{bloisi2011automatic,van2008discriminating,voles2000nautical}. In maritime scenario, horizon often appears as a linear feature and thus expected to be simple. However, several challenges may be encountered. An obvious challenge is the presence of the linear features of landmass, water-borne vessels, and air-borne objects. The other challenges encountered in visible range and near infrared sensors is the water patterns. Another challenge in data acquired from far infrared sensors is the continuous variation of intensity perpendicular to the horizon, which does not exhibit a sharp edge-like behaviour.

Conventional horizon detection methods use linear feature detectors such as Hough or Radon transform \cite{bloisi2011automatic,tang2013research,bao2005vision,zhang2010robusthorizon,wei2009automated,dusha2007attitude,mcgee2005obstacle}, statistical distributions representing the colors or intensity of the two regions created by a candidate horizon \cite{ettinger2003vision,fefilatyev2012detection,van2009polynomial,demonceaux2006omnidirectional,todorovic2004vision_ieee}, classification of pixels as belonging to sea and sky \cite{todorovic2004vision,bhanu1990model,ettinger2003vision}, trends of change of intensity close to the horizon \cite{bouma2008automatic,gershikov2013horizon}, and the use of statistical features \cite{fefilatyev2006horizon,van2009polynomial,demonceaux2006omnidirectional,bhanu1990model}. We note that more than one concepts have been used before. For example, Hough transform and distances between statistical distributions have been used in \cite{fefilatyev2012detection}. Multi-scale filtering has been used with intensity gradients in \cite{bouma2008automatic} and Radon transform in \cite{Prasad2017Muscowert}.

\begin{figure}
  \centering
  \includegraphics[width=0.82\linewidth]{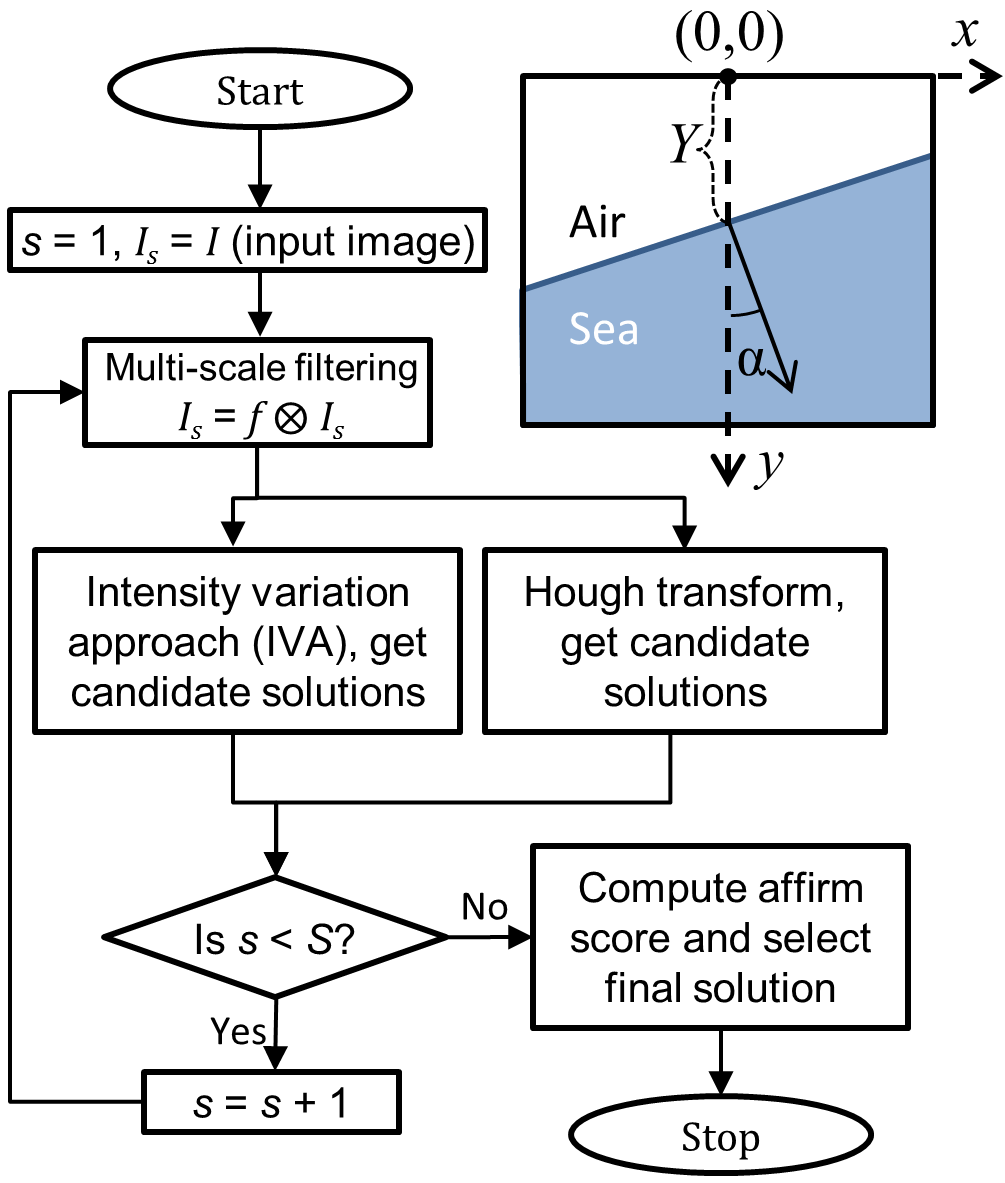}\\
  \caption{Flowchart of the proposed method.}\label{fig:flowchart}
\end{figure}

Linear feature detectors require that the line feature corresponding to horizon is represented by more number of edge pixels than any other line feature in the edge map. This requirement may not be met for horizon occluded by maritime objects or for presence of a long linear feature in image due to oceanic color differences, wakes, or long vessels in foreground \cite{prasad2017challenges}. Methods that use statistical distribution or statistical features are challenged by the color gamut of maritime videos which is largely the same in sky and sea regions but may have significant difference in the oceanic color due to depths and shadows. Methods that use trends of intensity change close to horizon are affected adversely by occlusion. A method that is robust to such scenarios is needed.

Here, we propose a new method called Multi-Scale Cross Modal Linear Feature (MSCM-LiFe). As the name implies, we use multi-scale approach. We additionally use two different modalities, Hough transform and intensity gradients, to find the line features which persist over these two modalities as well as over different scales. Thus, the method uses complementary as well as supplementary aspects of these approaches to provide better robustness to the above mentioned scenarios. The method is detailed in section \ref{sec:method}. Results and comparison with other methods are presented in \ref{sec:results}. The paper is concluded in section \ref{sec:conclusion}.

\section{Proposed method}\label{sec:method}

The flowchart of the proposed method is shown in Fig. \ref{fig:flowchart}. The horizon is represented using $Y$, the vertical distance of the horizon from the center point of upper edge of the image, and $\alpha$, the angle made by the normal to the horizon pointing into the sea with the $y-$ axis shown in inset of Fig. \ref{fig:flowchart}. The details of the functional blocks, namely multi-scale filtering, Hough transform, intensity variation approach (IVA), and the computation of affirm score for selection of the final solution are presented below.

\vspace{2mm}
\noindent \textbf{Multiscale filter \textemdash}~
Here we explain the computation of multi-scale images ${I_s}$ and mean multi-scale images $\tilde I_s$ which are used in the Hough transform block and IVA block, respectively.

\begin{figure*}
  \centering
  \includegraphics[width=0.95\linewidth]{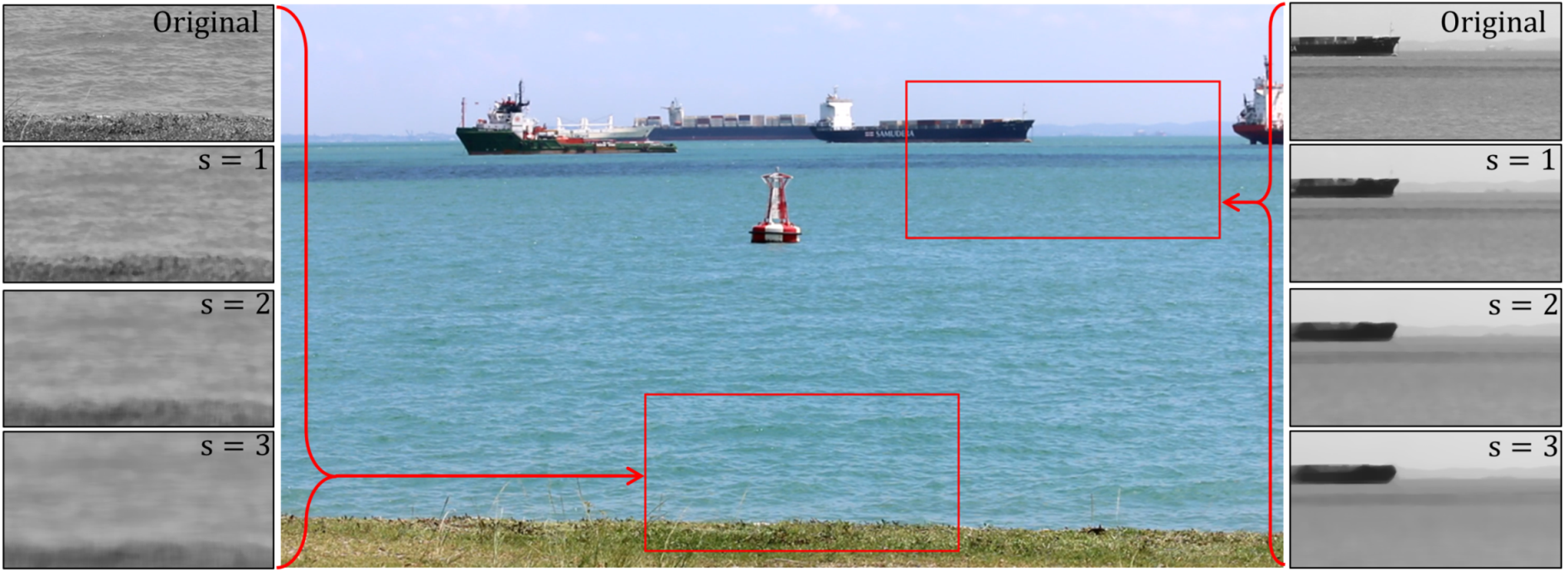}\\
  \caption{An example of the multi-scale images $I_s$ obtained from intensity image $I$ of the frame shown in the center}\label{fig:median}
\end{figure*}

For the input image $I$, we compute filtered image with vertical median filter of scale $s$ as follows:
\begin{equation}\label{eq:filter}
  {I_s}(x',y') = {f_s} \otimes I(x,y) = \mathop {{\rm{median}}}\limits_{(y' - y) \in [ - 2s,2s]} I(x',y')
\end{equation}
Such multi-scale images are are computed for $s=0\,\,{\rm to}\,\,10$. The median filter smooths small random variations but retains the prominent edges, thus aiding the Hough transform. An example of the multi-scale images $I_s$ obtained from an image $I$ is given in Fig. \ref{fig:median}.

Mean multi-scale image $\tilde I_s$ is also computed for each $s$ as follows:
\begin{equation}\label{eq:filter_mean}
  \tilde I_s(x,y) = \frac{1}{s} \sum_{s' \le s} I_s(x,y)
\end{equation}
The mean multi-scale image reinforces edges consistent over multiple scale, yet retains small details albeit with much smaller intensity. Bouma et. al reinforces used only the mean multi-scale image corresponding to the largest value of $s$. But, we use mean multi-scale image corresponding to each value of $s$ in the IVA block.

\vspace{2mm}
\noindent \textbf{Hough transform candidates \textemdash}~ First, the binary edgemap $E_s$ of the multi-scale image $I_s$ is computed using canny edge detector. Then, Hough transform is applied on the edge map $E_s$ as follows:
\begin{equation}\label{eq:Hough}
  H_s(\rho,\theta)=\int\limits_x { \int\limits_y{ E_s(x,y) \delta \big({ x \cos\theta + y\sin\theta - \rho}\big) dx dy}}
\end{equation}
The top 10 candidates with the largest values of $H_s(\rho,\theta)$ are selected for each scale $s$. Thus, a total of 100 candidates are collected using Hough transform. For each candidate $n$, the parameters of the horizon $C_n=(Y_n,\alpha_n)$ and the Hough score $H_n$ is stored.

\vspace{2mm}
\noindent \textbf{Intensity variation approach \textemdash}~ For each vertical strip (i.e. a column of pixels in an image) characterized by pixel $x$, the point $(x,y'(x))$ is determined as follows:
\begin{equation}\label{eq:IVA}
  y'(x)=\arg \max \limits_y \left|{\frac{{\rm d}\tilde I_s(x,y)}{{\rm d}y}}\right|
\end{equation}
This means that the point of maximum intensity variation is chosen for each strip of one pixel. Then, a line is fit on all these points. The line is used as a horizon candidate and $(Y,\alpha)$ corresponding to it is determined. For each scale $s$, the candidate $C_s=(Y_s,\alpha_s)$ is determined. The mean value of the magnitude of intensity gradients for all the strips is used as the IVA score:
\begin{equation}\label{eq:IVA_score}
  S_s= \mathop {{\rm{mean}}}\limits_x {\left| {\frac{{{\rm{d}}{{\tilde I}_s}(x,y)}}{{{\rm{d}}y}}} \right|_{y = y'(x)}}
\end{equation}

\vspace{2mm}
\noindent \textbf{Affirm score and selection of final solution}~ For each pair of Hough and IVA candidate $(n,s)$, we compute two scores, a goodness score $G(n,s)$ of the pair and the geometric proximity score $P(n,s)$ of the pair. The goodness score $G(n,s)$ quantifies the goodness of the pair in terms of the goodness of each candidate in the pair, i.e. goodness of the Hough transform for candidate $n$ and goodness of IVA for the candidate $s$. The goodness score $G(n,s)$ is defined as:
\begin{equation}\label{eq:Goodness}
  G(n,s)=H_n S_s
\end{equation}
The geometric proximity score $P(n,s)$ quantifies the geometric similarity of the candidates $n$ and $s$ in the candidate pair $(n,s)$. It is defined as follows:
\begin{equation}\label{eq:Proximity}
  P(n,s)=\left({1-\left({\frac{Y_n-Y_s}{\max(y)}}\right)^2}\right)\cos^2(\alpha_n-\alpha_s)
\end{equation}
where $\max(y)$ is the number of pixels along the y-axis (i.e., the number of pixels in the frame along the vertical direction, see inset of \ref{fig:flowchart}), ${(Y_n-Y_s)}/{\max(y)}$ is the relative vertical distance between horizon candidates $(n,s)$ and ranges between $[0,1]$. The normalization by ${\max(y)}$ is done because both Hough transform and IVA approaches gives candidates with the value of $Y$ in the range $[0,{\max(y)}]$. Consequently, $1-{(Y_n-Y_s)}/{\max(y)}$ represents the proximity of the candidates in terms of their vertical positions. The term $(\alpha_n-\alpha_s)$ represents the angular difference and lies in the range $[0,2\pi]$. However values close to $0$, $\pi$, and $2\pi$ mean that the candidates $(n,s)$ have similar orientation. Thus, $\cos^2(\alpha_n-\alpha_s)$ effectively represents proximity of the candidates $(n,s)$ in terms of orientation. Consequently, the geometric proximity score represents the proximity of the candidates $(n,s)$ in terms of both position and orientation.

Then, the affirm score is computed as the product of the goodness and geometric proximity scores:
\begin{equation}\label{eq:Affirm}
  A(n,s)=G(n,s) P(n,s)
\end{equation}
The pair with the highest value of the affirm score is used to determine the final solution. The high value of the affirm score indicates the best cross-modal pair, in which the Hough candidate corresponds to a true line feature. On the contrary, IVA candidates is subject to line fitting errors due to fitting of line over points determined independently for each vertical strip. Thus, instead of using a combination of both the Hough and IVA candidates, only the Hough candidate is used as the final solution.

\vspace{3mm}
\section{Results and comparison}\label{sec:results}
We present results of MSCM-LiFe and comparison with other methods in this section. We use 13 videos acquired using two Canon 70D cameras at 25 frames per second, one untouched camera for visible range videos, and another with hot mirror removed and NIR filter (BP800 Bandpass Filter by Midwest Optical Systems) added for near infrared videos. Among the visible range videos, 4 videos are taken on-board a ship while the other 5 videos are taken on-shore. Four NIR videos are taken in on-shore setup. Across these 13 videos, more that 3100 frames with various challenges are included. All frames are of high definition 1920 $\times$ 1080 pixels. Each frame in each video is annotated manually using a Matlab annotation tool and the annotation is used as the ground truth (GT). Across the $3100$ frames, the GT values of $Y$ range in $[215,905]$ and $\alpha$ range in $[-27.3^\circ,8.0^\circ]$. \textcolor{white}{\ref{tab:resultsSingapore}} 

\begin{table*}
  \centering
   \caption{Results for the videos in Singapore-Maritime-Dataset} \label{tab:resultsSingapore}
  \begin{tabular}{|l|r||r|r|r|r||r|r|r|r|}
     \hline
     {} & {} & \multicolumn{4}{|c|}{median($|Y-Y_{\rm GT}|$)} & \multicolumn{4}{|c|}{median($|\alpha-\alpha_{GT}|$) (degrees)}\\
     \cline{3-10}
     \hspace{-1mm}Video& \hspace{-1mm}Frames & MSCM-LiFe & ENIW \cite{ettinger2003vision} & FGSL \cite{fefilatyev2012detection} & MuSMF \cite{bouma2008automatic} & MSCM-LiFe & ENIW \cite{ettinger2003vision} & FGSL \cite{fefilatyev2012detection} & MuSMF \cite{bouma2008automatic} \\
      \hline
      \hline
      \multicolumn{10}{|c|}{Visible range on-board videos (camera is mounted on the ship and the ship is moving)}\\
      \hline
      V-1& 299 & {5.7} & { 37.8} & {37.5} & {170.2} & {0.6} & {0.5} & {0.5} & {0.5} \\
      V-2& 249 & {4.5} & {152.0} & {152.6} & {119.8} & {0.5} & {1.9} & {2.0} & {2.0}\\
      V-3& 249 & {3.2} & {174.8} & {17174.8} & {203.4} & {0.5} & {1.6} & {2.0} & {2.0} \\
      V-4& 249 & {6.5} & {69.5} & {69.8} & {331.2} & {0.7} & {1.2} & {1.3} & {1.3} \\
      \hline
      \hline
      \multicolumn{10}{|c|}{Visible range on-shore videos (camera is placed at shore on a fixed platform)}\\
      \hline
      V-5& {110} & {8.9} & {130.3} & {130.3} & {90.6} & {0.3} & {11.2} & {11.1} & {10.7}\\
      V-6 & {110} & {2.2} & {276.0} & {269.2} & {196.8} & {0.4} & {4.5} & {1.6} & {1.6} \\
      V-7 & {299} & {9.7} & {34.6} & {34.6} & {145.5} & {0.7} & {0.8} & {0.7} & {0.7} \\
      V-8 & {130} & {1.9} & {22.5} & {1.6} & {149.8} & {0.4} & {0.4} & {0.4} & {0.4}\\
      V-9 & {299} & {3.0} & {24.8} & {2.8} & {103.2} & {0.4} & {1.3} & {0.4} & {0.4}\\
      \hline
      \hline
      \multicolumn{10}{|c|}{Near-infrared on-shore videos (camera is placed at shore on a fixed platform)}\\
      \hline
      V-10 & {83} & {1.9} & {720.5} & {651.1} & {507.7} & {1.0} & {1.5} & {0.9} & {0.9}\\
      V-11 & {299} & {1.5} & {66.9} & {69.0} & {35.6} & {1.0} & {0.1} & {0.1} & {0.1} \\
      V-12 & {299} & {1.2} & {5.4} & {5.2} & {60.9} & {0.2} & {1.0} & {0.4} & {0.4}\\
      V-13 & {299} & {0.8} & {2.4} & {2.3} & {52.2} & {0.0} & {0.0} & {0.0} & {0.0}\\
      \hline
     Total & 3124 & 2.8 & 39.1 & 37.1 & 107.0 & 0.5 & 0.8& 0.5 & 2.0\\
     \hline
   \end{tabular}
   \vspace{2mm}
\end{table*}

We compare the performance of the proposed method with 3 other methods (Author's own Matlab implementations). For each method, the parameters $Y$ and $\alpha$ of the line detected as horizon are computed and compared with the ground truth. The methods used are the method of Ettinger et. al (ENIW) \cite{ettinger2003vision}, method of Fefilatyev et. al (FGSL) \cite{fefilatyev2012detection}, and multi-scale median filter based approach (MuSMF) \cite{bouma2008automatic}. These methods represent the state-of-the-art for horizon detection in maritime images. Further, Hough transform is used in FGSL and multi-scale IVA is used MuSMF. Thus, they can be considered predecessors of the proposed method.

\begin{figure*}
  \centering
  \begin{tabular}{cc}
    \hspace{-2mm}\includegraphics[width=0.45\linewidth]{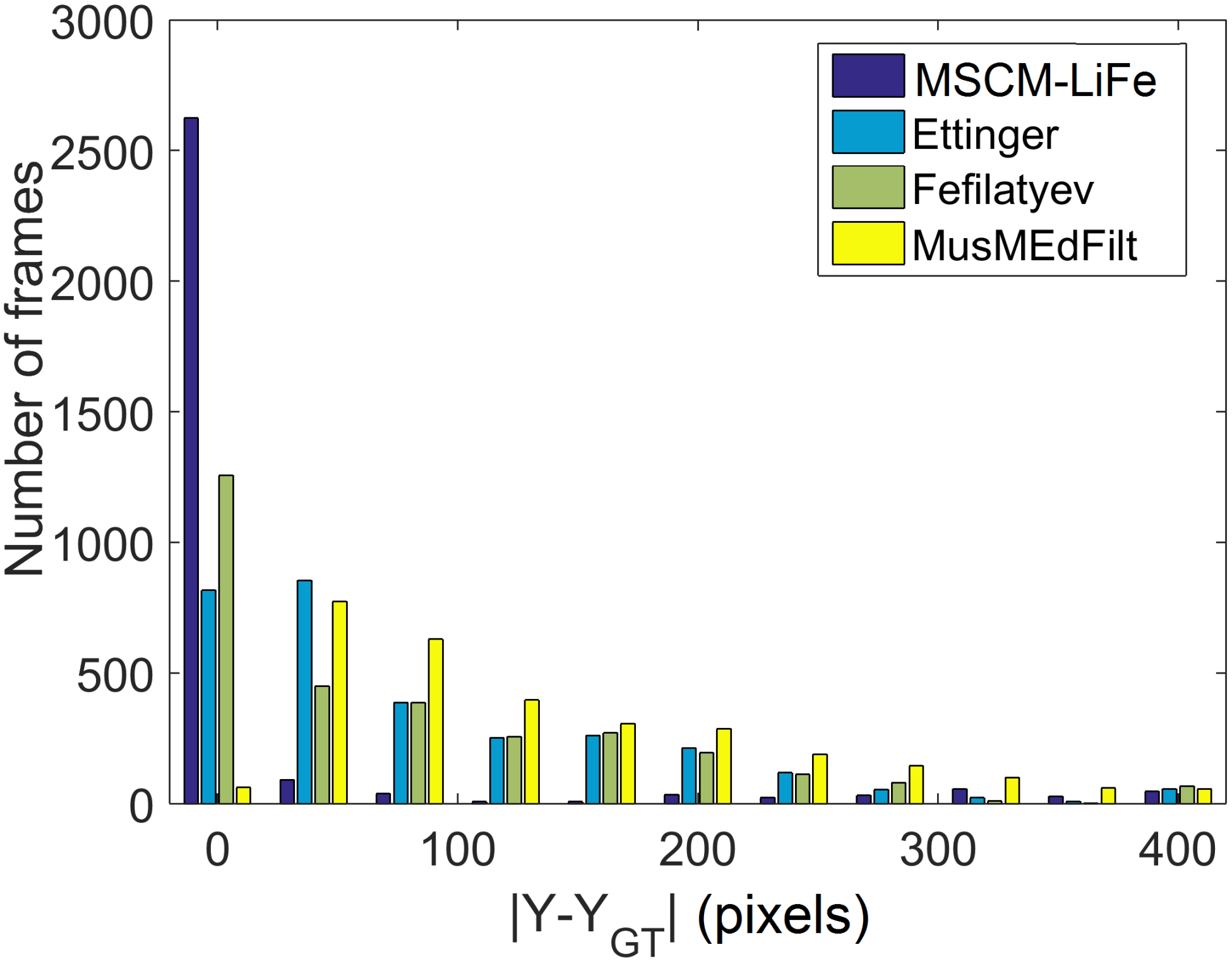} & \hspace{-2mm}\includegraphics[width=0.45\linewidth]{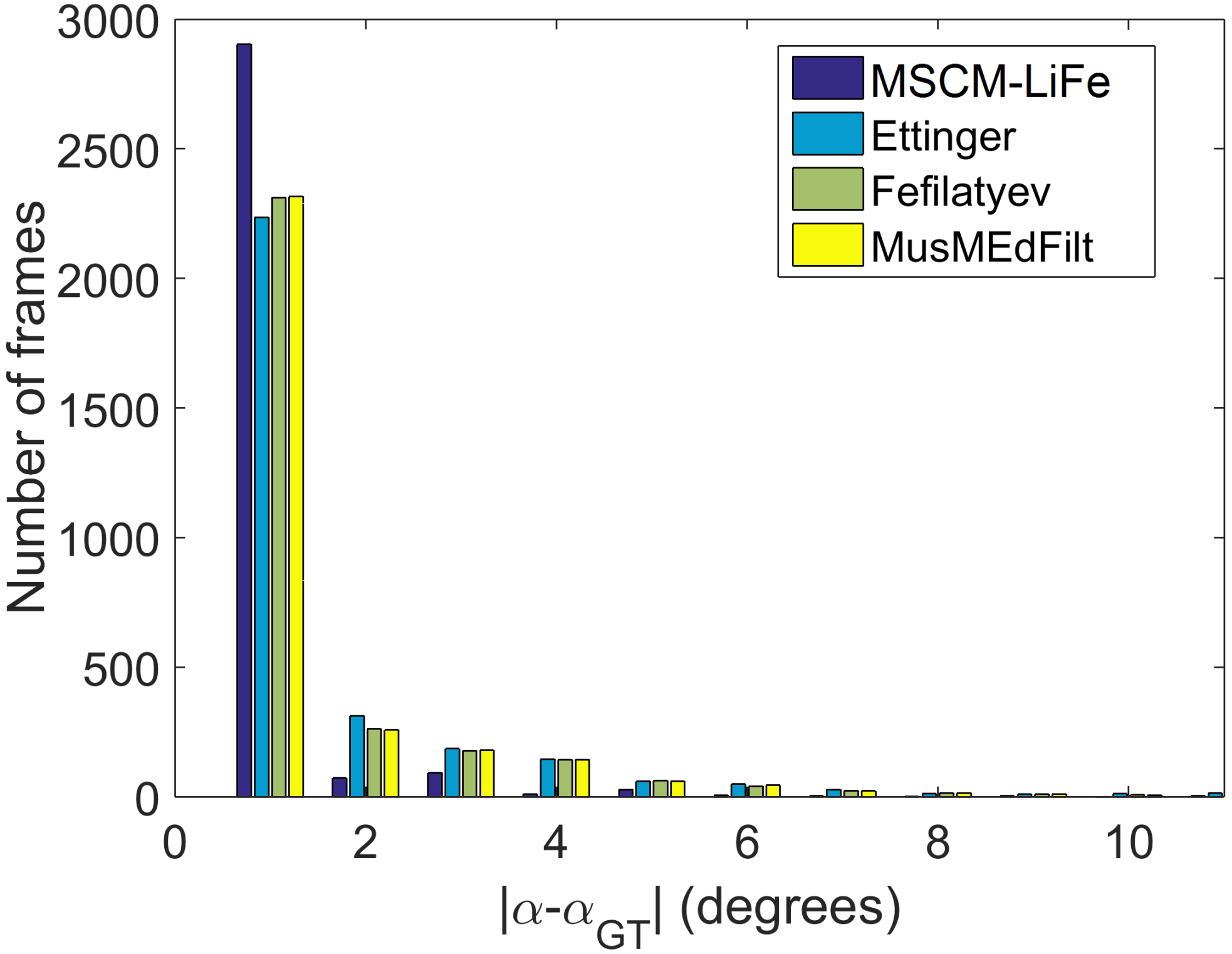} \\
    (a) Histogram of error in estimate of $Y$ for all methods & (b) Histogram of error in estimate of $\alpha$ for all methods\\
    \hspace{-3mm}\includegraphics[width=0.45\linewidth]{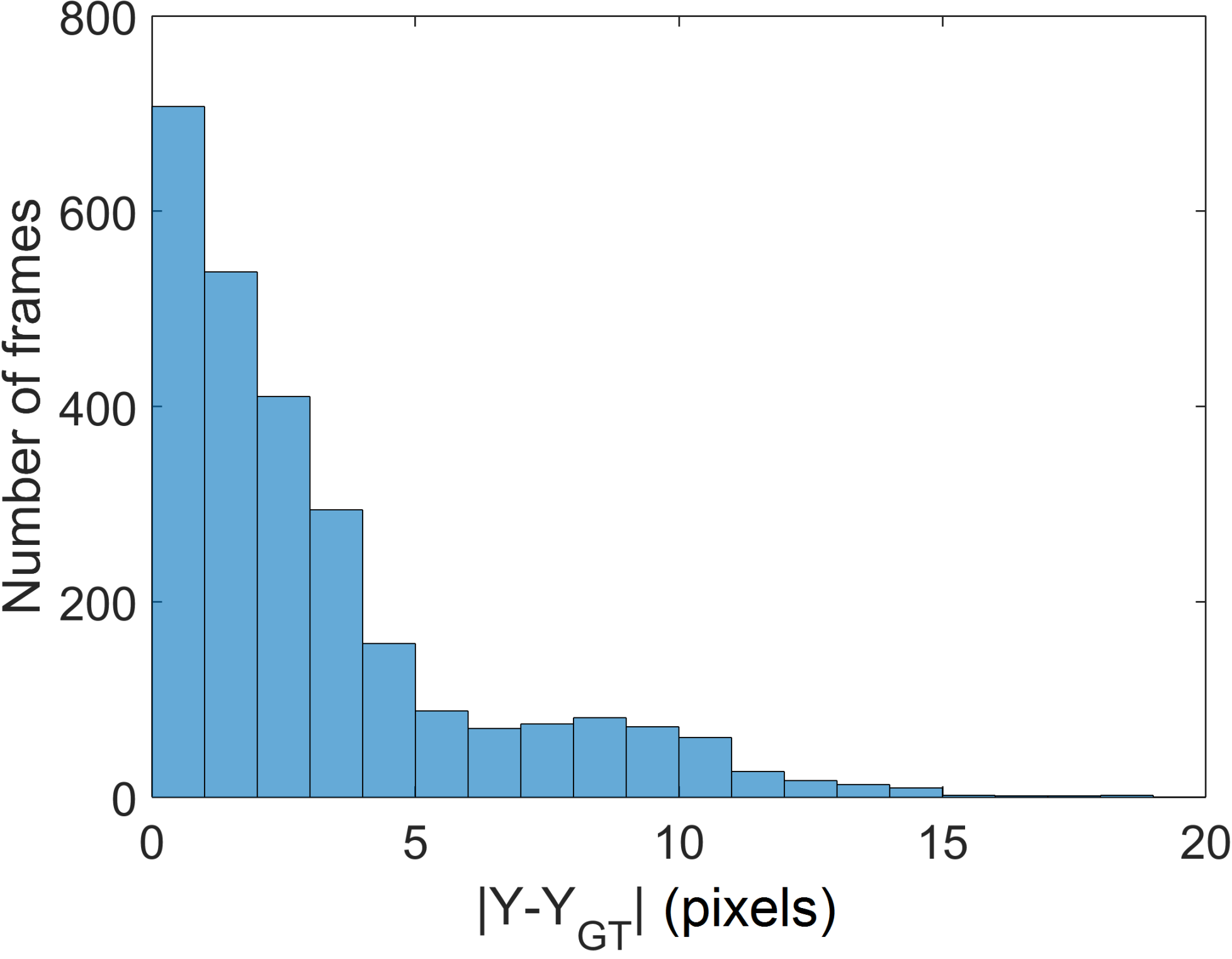} & \hspace{-3mm}\includegraphics[width=0.45\linewidth]{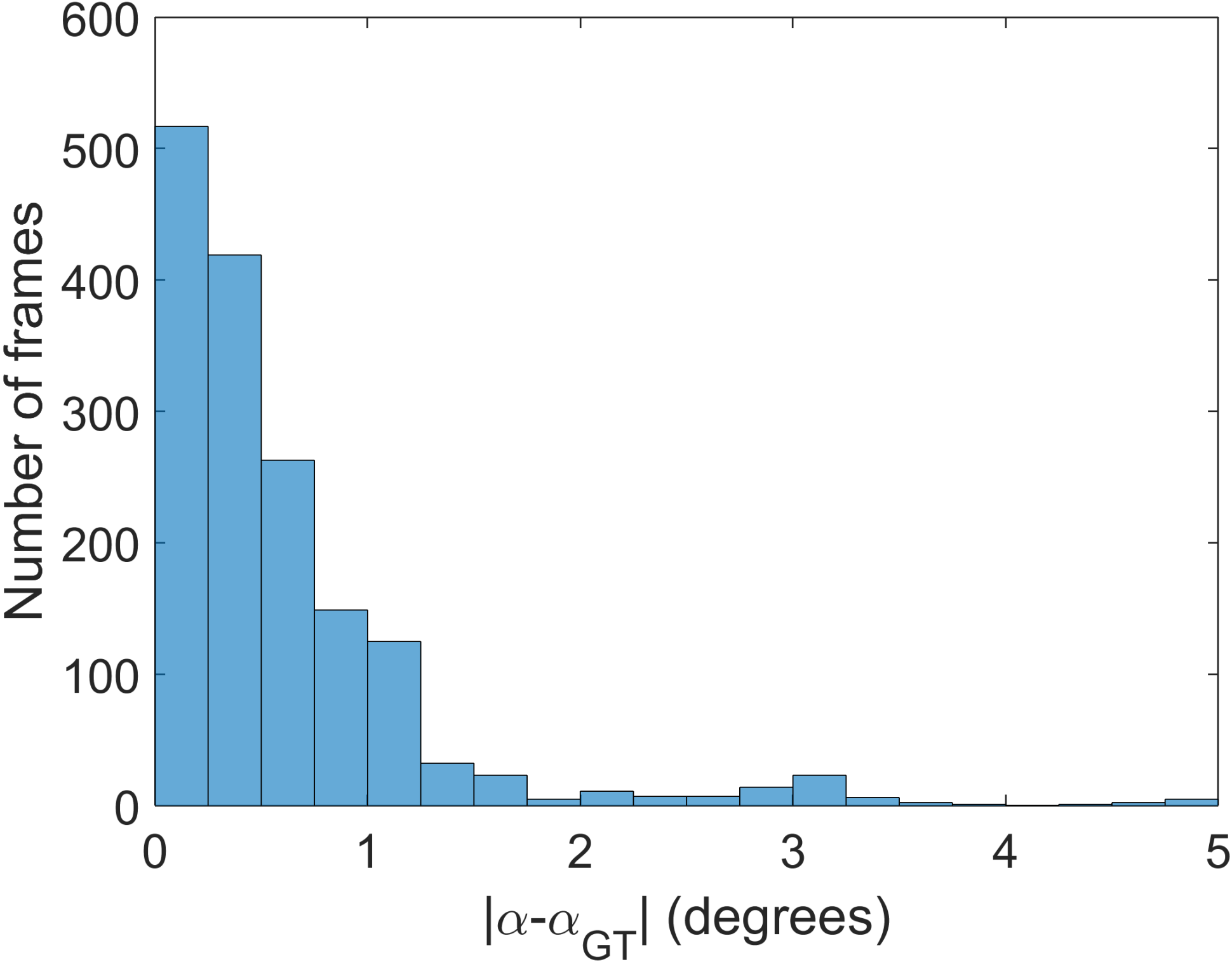} \\
    (c) Histogram zoom-in for MSCM-LiFe for $Y$ & (d) Histogram zoom-in for MSCM-LiFe for $\alpha$\\
  \end{tabular}
\caption{Statistical distribution of errors of the various methods (a,b) and finer distribution for MSCM-LIFe (c,d)}\label{fig:results}
\end{figure*}

The median of errors in the parameters $Y,\alpha$ for each of the videos are listed in Table \ref{tab:resultsSingapore}. It is seen that MSCM-LiFe performs consistently well giving a small median error for all of the videos. In comparison, all other methods perform poorer. The error is more accentuated in the parameter $Y$. Only for video V-13, the errors of 3 methods is parameter $Y$ is less than 2.5 pixels.

\begin{figure*}
  \centering
  \includegraphics[width=1\linewidth]{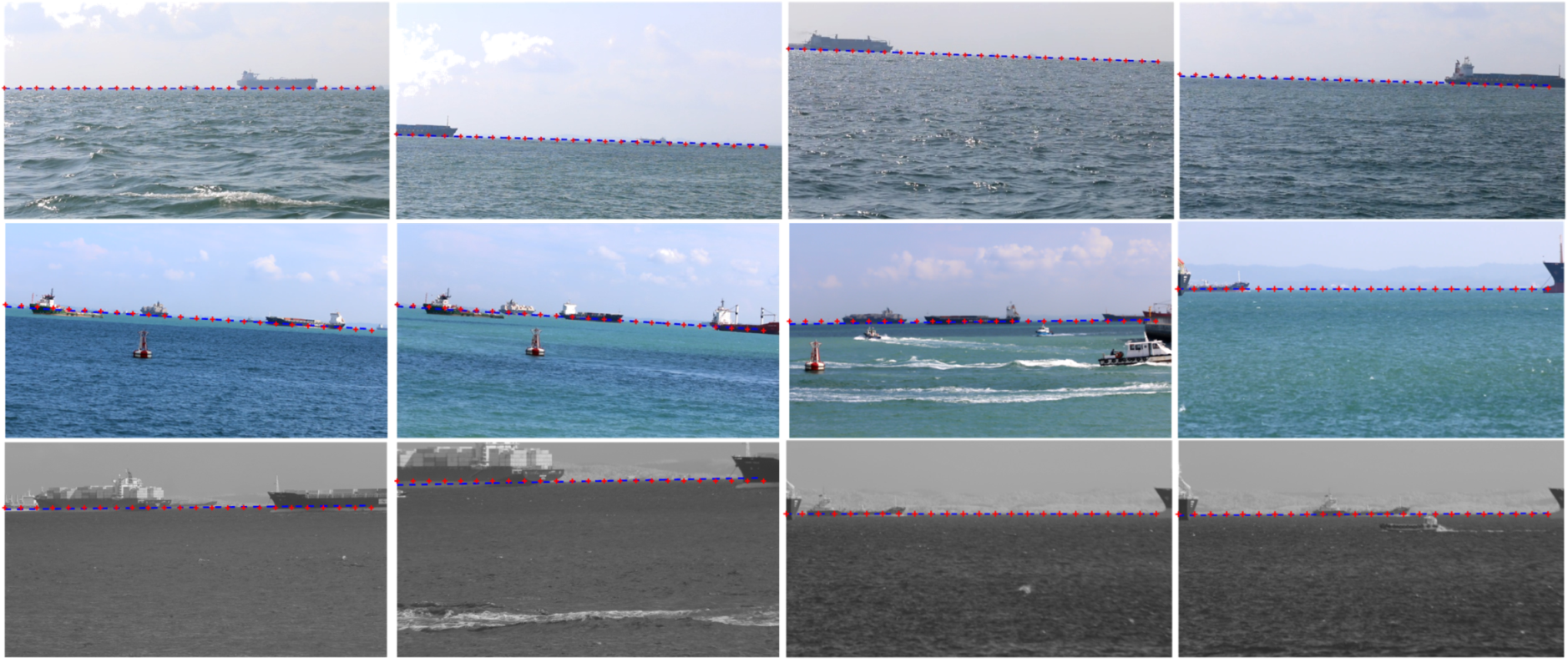}\\
  \caption{Sample frames are shown here, dashed blue line show ground truth and red markers show the result of the proposed method (color available online only). Gray color images are from NIR videos.}\label{fig:sample}
\end{figure*}

We also plot the error histograms for both $Y$ and $\alpha$ parameters for all the frames in Fig. \ref{fig:results}(a,b). While all algorithms have peak error values in the first bin, which is of the size 40 pixels for $Y$ and 5$^\circ$ for $\alpha$, the proposed method have highest number of frames with peak error in the first bin which means lower error comparatively. A further zoom-in of the histograms for the proposed methods is shown in Fig. \ref{fig:results}(c,d), which clearly indicates that the proposed method has very small error for most of the frames, within 10 pixels for the parameter $Y$ for 80\% of the frames and within 1$^\circ$ for the parameter $\alpha$ for 86\% of the frames. Some sample results are shown in Fig. \ref{fig:sample}.

The small percent of frames for which the MSCM-LiFe performs poor are typically the frames in which the foam of a long wake spanning the entire width of the frame causes very strong Hough candidates as well IVA candidates which are consistent across all the scales. Notably, all the methods fail in such scenario.

\section{Conclusion}\label{sec:conclusion}

This paper has proposed multi-scale cross-modal line feature for horizon detection to overcome the challenges of horizon detection in images captured using electro-optical sensors. Usage of three concepts and two modalities ensures that the line feature persistent over multiple scales and occurring across more than one modality is chosen. The choice of such line feature is dictated by a newly proposed affirm score that checks the goodness of a line feature pair across the two modalities as well as the geometric proximity of the pair.

The method is tested for 13 videos comprising of more than 3000 frames, of which 4 videos were near-infrared and the others were visible range. The results show that the proposed method performs well even in the presence of severe occlusion, wakes, skyline, etc. It performs better than 3 contemporary methods which use concepts such as mutli-scale filtering, Hough transform, intensity variation, as well as statistical distances.

\section*{Acknowledgment}
This work was conducted within Rolls Royce@NTU Corporate Lab with the support of National Research Foundation under the CorpLab@University scheme.

\bibliographystyle{IEEEtran}

\end{document}